\documentclass[conference]{IEEEtran}
\IEEEoverridecommandlockouts
\usepackage{cite}
\usepackage{amsmath,amssymb,amsfonts}
\usepackage{algorithmic}
\usepackage{graphicx}
\usepackage{textcomp}
\usepackage{xcolor}
\usepackage{tabularx}
\usepackage{hyperref}
\usepackage{multirow}

\usepackage{siunitx}

\usepackage[capitalise]{cleveref}
\usepackage{flushend}

\def\BibTeX{{\rm B\kern-.05em{\sc i\kern-.025em b}\kern-.08em
    T\kern-.1667em\lower.7ex\hbox{E}\kern-.125emX}}
\begin{document}

\title{
Two-stage Early Prediction Framework of Remaining Useful Life for Lithium-ion Batteries
}

\author{\IEEEauthorblockN{
            Dhruv Aditya Mittal\IEEEauthorrefmark{1},
            Hymalai Bello\IEEEauthorrefmark{2},
            Bo Zhou\IEEEauthorrefmark{1}\IEEEauthorrefmark{2},
            Mayank Shekhar Jha\IEEEauthorrefmark{3},
            Sungho Suh\IEEEauthorrefmark{1}\IEEEauthorrefmark{2}\thanks{Corresponding author: sungho.suh@dfki.de},
            Paul Lukowicz\IEEEauthorrefmark{1}\IEEEauthorrefmark{2}}
        \IEEEauthorblockA{\IEEEauthorrefmark{1}Department of Computer Science, RPTU Kaiserslautern-Landau, Germany}
        \IEEEauthorblockA{\IEEEauthorrefmark{2}German Research Center for Artificial Intelligence (DFKI), Kaiserslautern, Germany}
        \IEEEauthorblockA{\IEEEauthorrefmark{3}Centre de Recherche en Automatique de Nancy (CRAN), University of Lorraine, France}}

\maketitle

\begin{abstract}
Early prediction of remaining useful life (RUL) is crucial for effective battery management across various industries, ranging from household appliances to large-scale applications. Accurate RUL prediction improves the reliability and maintainability of battery technology. However, existing methods have limitations, including assumptions of data from the same sensors or distribution, foreknowledge of the end of life (EOL), and neglect to determine the first prediction cycle (FPC) to identify the start of the unhealthy stage. This paper proposes a novel method for RUL prediction of Lithium-ion batteries. The proposed framework comprises two stages: determining the FPC using a neural network-based model to divide the degradation data into distinct health states and predicting the degradation pattern after the FPC to estimate the remaining useful life as a percentage. Experimental results demonstrate that the proposed method outperforms conventional approaches in terms of RUL prediction. Furthermore, the proposed method shows promise for real-world scenarios, providing improved accuracy and applicability for battery management.

\end{abstract}

\begin{IEEEkeywords}
 Remaining Useful Life Prediction, Lithium-ion Batteries, Degradation Modelling
\end{IEEEkeywords}

\section{Introduction}

Lithium-ion batteries are widely used in various industries due to their high energy density and versatility. Some broader applications are consumer electronics, electric vehicles, medical devices, aerospace and defense, and renewable energy storage \cite{dos2021lithium}. With its increasing demand in many safety-critical applications, it becomes essential to predict the life of the batteries to avoid any adverse scenarios. Accurate prediction of these batteries' remaining useful life (RUL) is crucial for efficient maintenance planning, risk management, and cost optimization. By identifying potential battery failures or degradation in advance, operators can take proactive measures to mitigate risks, avoid unexpected downtime, and optimize resource allocation.

Researchers have proposed many methods to predict the RUL \cite{salkind1999determination,miao2013remaining}.
These approaches are divided into Data-based, Model-based, and Hybrid Approaches.
Model-based methods require a physical degradation model to approximate the Li-ion battery's internal electrochemical reactions using mathematical equations.
For example, the deduction of the mathematical model can employ an equivalent circuit to identify the internal resistance influencing the degradation data \cite{remmlinger2011state}. 
On the one hand, using a mathematical model is a memory, latency, and energy-efficient solution. 
In addition, it offers robust validation and verification of the model before deployment. 
On the other hand, in Model-based approaches, the parameter selection and initial conditions play an essential role. 
Slight deviations from the optimized values can lead to significant inaccuracies in RUL predictions. 
Also, the precision of the mathematical/physical model is usually negatively impacted by non-linearities in the system. 
Data-driven approaches offer the advantages of reducing parameter dependency and learning hidden patterns in non-linear data. 
Moreover, Data-based methods require minimal expert knowledge to develop a robust solution. 
The Hybrid approach has been proposed to exploit the advantages of Model-based and data-driven solutions. 
The Hybrid methodology combines the knowledge of causality from historical battery degradation data with applicable physics-based models such as; chemical reactions and heat transfer that govern the battery behavior \cite{wu2021hybrid, li2019remaining}. 

However, conventional RUL prediction methods have certain limitations. For example, the conventional deep learning-based models have tried to input some percentage of the initial discharge capacity to get the discharge values for later timestamps/cycles. In addition, the above-mentioned data-driven methods assume that training and test data are collected by the same sensors under the same operating conditions or are from the same distribution. However, these assumptions can be impractical in real-world scenarios, as battery working conditions often vary depending on specific tasks and different data sources. Consequently, these methods may struggle to adapt to the changing operating conditions and accurately predict RUL in diverse environments. 

Moreover, accurately determining a battery's health state (HS) is essential for precise RUL prediction. Unfortunately, conventional hybrid methods often neglect to identify the first prediction cycle (FPC), which marks the beginning of the unhealthy stage \cite{suh2020supervised, suh2022generalized}. Failure to determine the FPC can lead to suboptimal RUL predictions, as the HS of the battery may exhibit little differences in the run-to-failure training dataset.

We propose a novel two-stage RUL prediction framework to overcome these limitations in this paper. The first stage focuses on predicting the FPC, enabling accurate identification of the unhealthy stage. By dividing the degradation data into distinct health stages, we ensure precise identification of the FPC. In the second stage, we predict the degradation pattern that occurs after the FPC, enabling the estimation of the RUL as a percentage. Furthermore, we incorporate multiple battery degradation features, including discharge capacity, charge capacity, temperature, internal resistance, and charge time. By considering a broader range of features, we improve our model's prediction accuracy and robustness. 

Experimental results demonstrate that the proposed framework outperforms conventional methods in terms of RUL prediction on the MIT dataset \cite{severson2019data}, consisting of 124 battery cells. The main contributions of our study can be summarized as follows:
\begin{itemize}
    \item We propose a novel two-stage framework for RUL prediction, where the first stage predicts the FPC and the second stage predicts the degradation pattern after the FPC.
    \item We annotate the degradation pattern automatically by determining the FPC to train the proposed neural network models.
    \item We incorporate multiple battery degradation features for RUL prediction, improving the prediction accuracy and robustness, while most conventional methods estimate RUL using only discharge capacity.
    \item Experimental results demonstrate the superior performance of the proposed method over conventional methods on the MIT dataset with 124 different battery cells.
\end{itemize}

The rest of the paper is organized as follows; \cref{sec:relatedwork} introduces the related works. \cref{sec:datsets} provides detailed information on the dataset we used. \cref{sec:method} introduces the details of the proposed two-stage framework for RUL prediction, and \cref{sec:experiments} presents the experimental results and analysis. Finally, \cref{sec:conclusion} concludes the paper.

\section{Related Work}
\label{sec:relatedwork}

In this section, we further discuss some of the existing data-driven models.
Deep learning techniques have confirmed their relevance in increasing the RUL estimation's accuracy, consequently improving the power of predictive energy management \cite{wang2021critical}. 
One of the most common deep learning approaches is based on recurrent neural networks and long short-term memory algorithms (LSTM) \cite{zhang2018long, hong2019synchronous, zhang2017lstm}. 
\cite{song2018lithium} proposed dynamic LSTM leading to online RUL prediction wherein indirect voltage measures were used for health index creation rather than the capacity of the battery, \cite{liu2019deep} proposed ensemble approach with LSTM wherein uncertainty was taken into account with bayesian model averaging, \cite{ren2021method} integrated LSTM with particle swarm optimization wherein the use of PSO leads to the efficient calculation of LSTM parameters matching the given topology.  LSTMS variants such as gated recurrent units (GRUs) have been extensively explored as well. Some notable works include: \cite{hannan2020state} that address the state of charge (SOC) estimation under variable ambient temperatures, \cite{huang2019convolutional} proposed convolutional GRUs, and \cite{cui2020state} that integrated and employed attention mechanism within GRUs for battery prognostics. 
On the other hand, the convolutional neural network (CNN) enables the integration of other methods for better performance. For instance, \cite{zhou2020state} employed causal convolution and dilated convolution techniques in the model to enhance its capability to capture and represent local contextual information.
\cite{yang2020remaining} combined CNNs and Bi-LSTM in such a way that CNN enabled efficient feature extraction and its feeding to Bi-LSTM. \cite{zhao2020lithium} proposed  RNN-CNN combination wherein RNN extracted the features to lessen time dependency followed by their injection to CNN, \cite{xu2021life} proposed a compact pruned CNN followed by incorporation of transfer learning for efficient battery capacity prediction.

Several researchers have employed extreme learning machine (ELM) due to its fast learning speed and generalization performance with a stable estimation of the RUL in Lithium-ion batteries \cite{wang2021critical, ma2020capacity, zhang2023indirect}. 
Variations of the ELM method have been proposed to remove the need to set the number of neurons and weights in the hidden layer, such as kernel ELM (KELM) \cite{heidari2019efficient}, multiple kernel ELM (MKELM) \cite{liu2022remaining} and the combination of MKELM with sparrow search algorithm (ISSA-MKELM) \cite{zhang2023indirect}. 
In \cite{wang2022online}, a bidirectional LSTM (Bi-LSTM) scheme in combination with an attention mechanism is adopted to estimate the state of health (SOH). 
The approach outperforms the vanilla version of LSTM and Bi-LSTM models.
\cite{tong2021early} combined the features of adaptive dropout long short-term memory (ADLSTM) and Monte Carlo (MC) simulation for RUL prediction. 
Their method achieves comparable accuracy with other peers' works, despite the use of only 25 \% of the discharge capacity data as input, compared with the counterparts that employed a range between 40 and 70 \% of the discharge capacity data as input.  
The proposed method outperforms the mainstream regression algorithms and several recently published hybrid methods regarding data requirement and prediction accuracy. Also, there are some multi-stage RUL prediction techniques, where \cite{ma2023two} employed CNN to extract the discharge capacity features and applied the Gaussian process regression (GPR) algorithm, whereas \cite{yao2022two} extracted features from multiple cycles to form a time series dataset and then applied different regression models for the lifetime prediction of lithium-ion batteries.



\section{Datasets}
\label{sec:datsets}

\begin{figure}[!t]
\includegraphics[width=\columnwidth]{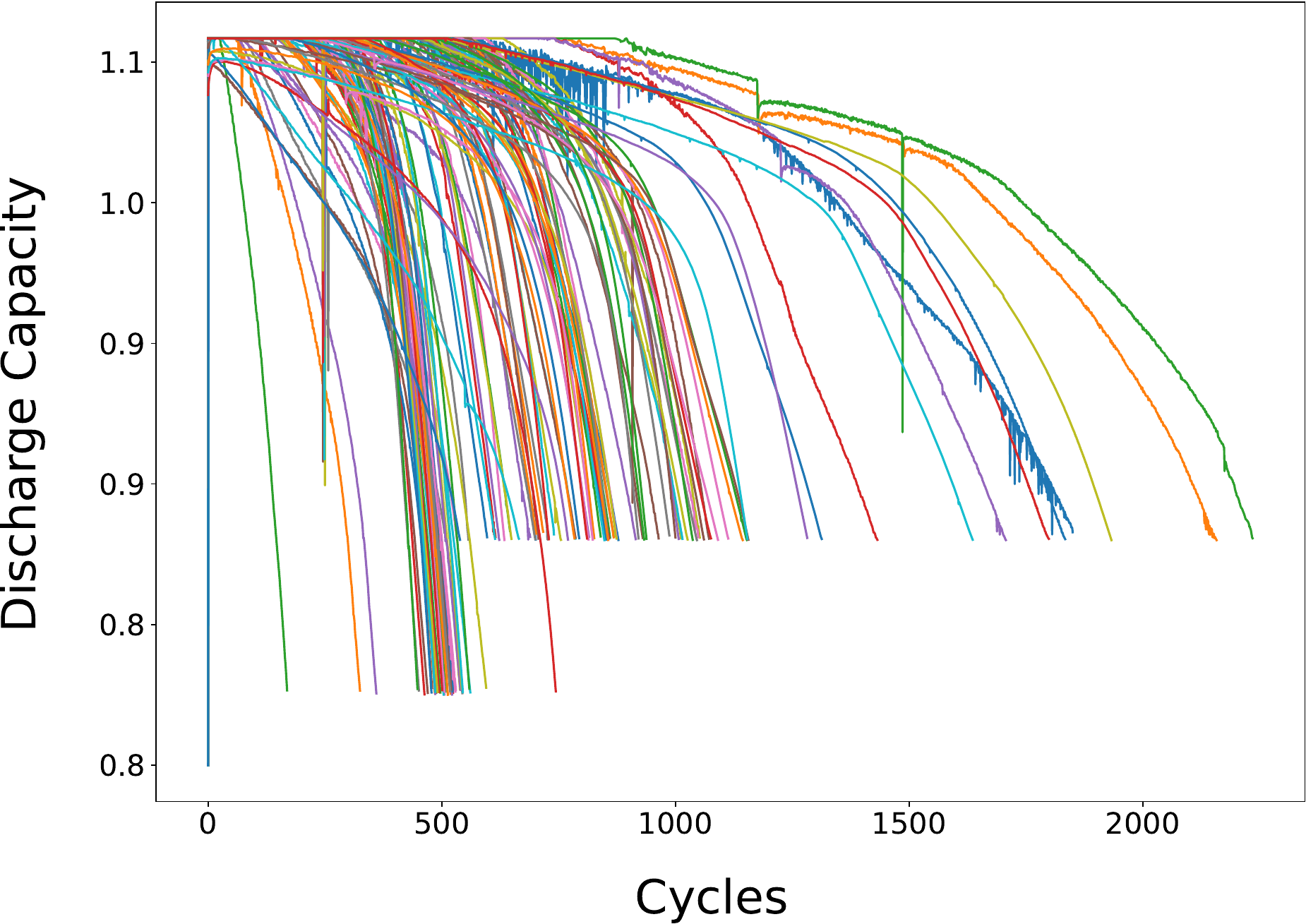}
\caption{Discharge Capacity Pattern with Number of Cycles for 124 Lithium-ion Batteries in the MIT Dataset}
\label{MIT_Data}
\end{figure}

\begin{figure}[!t]
\includegraphics[width=\columnwidth]{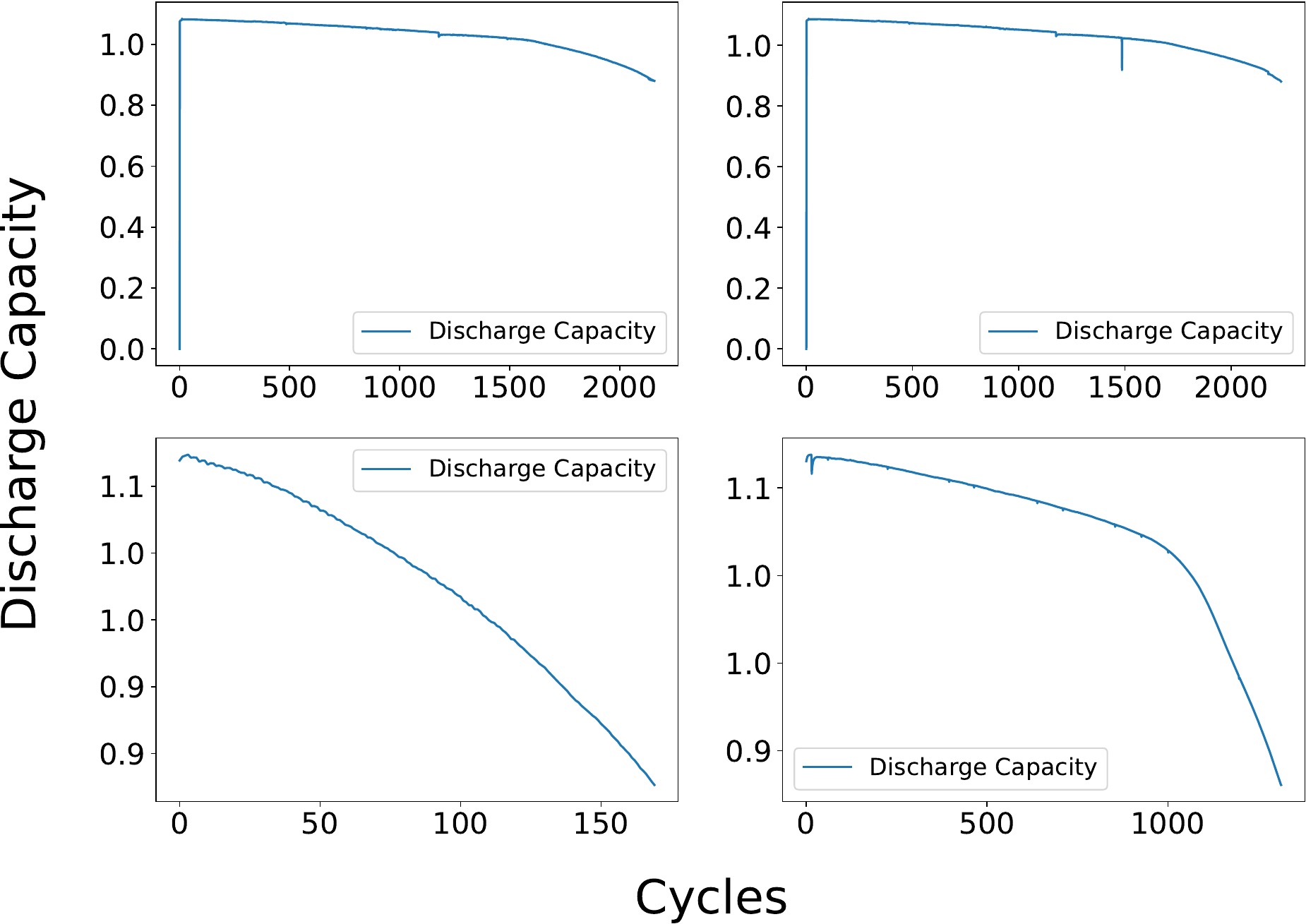}
\caption{Discharge Capacity with Completed Cycles of Four Different Batteries in the MIT Dataset. \textbf{First Row} Cells One and Two with Cycles $\geq$ 2000. \textbf{Second Row}  Cell Forty Two with $\le$ 175 Cycles, and Cell One Hundred with $\leq$ 1300 Cycles}
\label{mit_dataset_zoomed}
\end{figure}

In this section, we discuss our dataset for the task of RUL prediction in Lithium-ion batteries: the MIT dataset \cite{severson2019data}. The MIT dataset, provided by A123 Systems, consists of 124 commercial lithium iron phosphate graphite cells (A123 Systems, A123 Systems, model APR18650M1A, 1.1 Ah nominal capacity) that were cycled under fast-charging conditions in a temperature-controlled environmental chamber (30 \si{\celsius}) \cite{severson2019data}. The dataset encompasses a wide range of cycle lives, varying from 150 to 2,300 cycles. Each cycle represents a complete discharge and recharge of the battery. The dataset includes various parameters such as cycles, internal resistance, minimum temperature, average temperature, maximum temperature, charge capacity, and discharge capacity. \cref{MIT_Data} illustrates the discharge capacity data of 124 cells, revealing the noticeable variance in the number of cycles among the Lithium-ion batteries. The dataset has been widely used for RUL prediction research \cite{sanz2021remaining, afshari2021remaining, wang2022online} and is available at \url{https://data.matr.io/1}



For our experimentation, we evaluate the proposed method on the MIT dataset due to its larger number of cells and the significant variance in the number of cycles, which allows for a more comprehensive analysis and prediction of patterns. \cref{mit_dataset_zoomed} shows the different battery degradation patterns in the MIT dataset and a closer look at the variance in the number of cycles in the dataset. The MIT dataset also records seven different channels mentioned above. To generate our dataset for training and evaluation, we employed a rolling window approach with a window size of 50 and a step value of 1.

\section{Methodology}
\label{sec:method}


\begin{figure*}[!t]

\centering
\includegraphics[width=\linewidth]{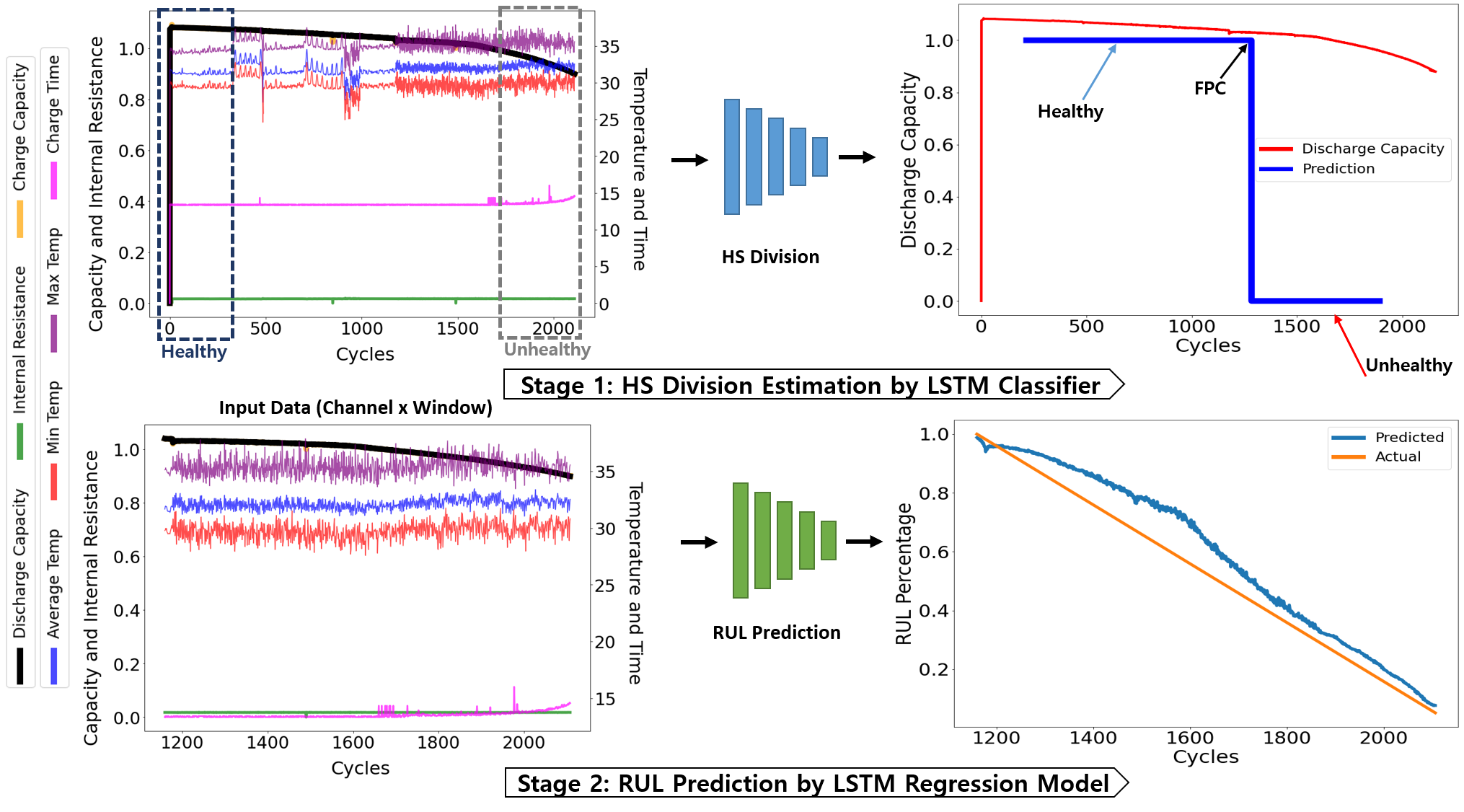}
\caption{Overview of The Proposed Framework. \textbf{First Row} Healthy State Estimation Step Using an LSTM Classifier (First Predicted Cycle Recognition). \textbf{Second Row} Remaining Useful Life of Batteries with the Use of LSTM Regression Model.}
\label{flow}
\end{figure*}

In this section, we present the proposed framework, which is illustrated in \cref{flow}. The proposed framework addresses the challenge of varying degradation patterns among different battery cells by incorporating a two-stage approach. Unlike conventional approaches, the proposed framework begins by classifying data of each cycle into healthy and unhealthy states, as this classification helps improve the RUL prediction. 
The proposed framework consists of two stages: the FPC stage and the RUL prediction stage. In the FPC decision stage, we aim to classify the healthy and unhealthy states of the battery cells. By separating these states, we can determine the FPC point, which marks the start of the unhealthy stage. This stage is crucial as accurately identifying the FPC enables accurate RUL prediction. After the FPC point is determined, we define the remaining data after the FPC as degradation data. Then, utilizing all available features, such as discharge capacity, charge capacity, charging time, temperature, and internal resistance, we predict the RUL in percentage from the FPC to the end-of-life (EOL) using the LSTM-based RUL prediction model.

\subsection{First Prediction Cycle Decision}

\begin{table}[!t]
\caption{\label{table:model_architecture}Network Architecture}
\centering
\renewcommand{\arraystretch}{1.3}
\resizebox{\linewidth}{!}{
\begin{tabular}{ | c|c|c|c|c|c| } 
\hline
Network & Name &Layers &Activation & Output Dimension \\ 
\hline
\multirow{5}{6em}{Feature Extractor}
    & Input    &   -                                    &  -     &  7x50 \\ 
    &LSTM1    & 4 x LSTM (hiddenSize = 50)              &Tanh  & 7x50\\ 
    &LSTM2    & 4 x LSTM (hiddenSize = 50)             &Tanh  & 7x50\\
    &Flatten  &   -                                     &   -    & 350\\ 
    &Dense1  &   FC (128)                           &   ReLU    & 128\\ 
\hline
HS Division & DenseSigmoid   &  FC (1)              &   Sigmoid  &  1\\
\hline
RUL Prediction & DenseRelu      &  FC (1)         &  ReLU  &  1 \\
\hline
\end{tabular}}
\end{table}

Predicting time-series data is challenging due to the limited availability of data and the non-linear nature of degradation patterns. Time-series data is influenced by various factors, such as usage patterns, load variations, and operating conditions, making it difficult to measure internal reactions within the battery. To overcome the limitation of data, we employed a sliding window with a size of 50 and a sliding step of 1, including 7 different channels mentioned in \cref{sec:datsets}.

In the first stage of our framework, the objective is to determine the FPC point, which indicates the transition from the healthy state to the unhealthy state of the battery. However, annotating the unhealthy and healthy parts of the data for training the FPC decision network model is challenging. We define the first 10\% of the cycles as the healthy state and the last 10\% of the cycles as the unhealthy state in the training data, as shown in \cref{flow}. This manner not only reduces the burden of manual labeling but also improves the prediction of the HS division when no ground truth of HS is available in most of the open-access Lithium-ion battery datasets. Let $X^i=[x_1^i,...x_n^i]$ be the input data with all features of $i$-th battery cell by applying a sliding window of size $n_w$ of the $n_f$ available features with $x_j^i \in \mathbb{R}^{n_f \times n_w}$ consisting of the battery cycle data for that window. The corresponding HS labels are defined as follows.
\begin{equation}
    \label{eq:HSLabel}
    {y_{HS}}_j^i=\left\{
	\begin{array}{@{}ll@{}}
	0, & \text{if}\ j<\text{EOL}^i\times p \\
	1, & \text{if}\ j>\text{EOL}^i\times (1-p)
	\end{array}\right.
\end{equation}
where ${y_{HS}}_j^i$ represents the HS label of $x^i_j$, EOL$^i$ denotes the total number of cycles in the $i$-th battery cell data, and $p$ is the percentage of the total degradation process which is set to 10\% in this work. Thus, a total of 20 \% of training data are labeled and utilized to train the HS division network model in the FPC decision stage. 

The HS division model utilizes a 1D LSTM architecture, as depicted in \cref{table:model_architecture}. The network model consists of two LSTM modules, each having 4 layers. Following the LSTM layers, we adopt two fully-connected (FC) layers to classify the HS, with the sigmoid activation function. To train the HS division model, we use the binary cross entropy (BCE) loss function between the predictions and the corresponding HS labels.
\begin{equation}
    \label{eq:BCELoss}
    \resizebox{0.91\columnwidth}{!}{
    $
    L_{BCE} = E_{{y_{HS}^i},{\hat{y}_{HS}}^i}[{y_{HS}^i}\log {\hat{y}_{HS}}^i + (1-{y_{HS}^i})\log (1-{\hat{y}_{HS}}^i)]
    $
    }
\end{equation}
where ${\hat{y}_{HS}}^i$ and ${y_{HS}^i}$ denote the HS division prediction and the corresponding HS label of the $i$-th battery cell data, respectively.

Once the HS division model is trained, it can classify the remaining training and test data into healthy and unhealthy states without specifying a threshold value. As the model learned differences in the features of the input data between the labeled healthy and unhealthy data during the training procedure, the model can recognize degradation patterns across the entire dataset. It is important to determine the FPC, which is a starting cycle of the deterioration of the Lithium-ion battery cell. However, it is challenging to determine the FPC, as the degradation features at this stage are often weaker than those in the labeled unhealthy stage. To determine appropriate FPC, we adopt a simple continuous trigger mechanism. In this mechanism, the battery cell is considered to be in an unhealthy state when a certain number of consecutive unhealthy predictions are provided by the HS division model. This trigger mechanism helps prevent unnecessary oscillation and uncertainty in determining the FPC at an earlier stage. The trigger mechanism is a common strategy in HS division approaches \cite{lei2018machinery, suh2020supervised}. In this paper, we determine the FPC if the HS division model classifies the input data as the unhealthy class 5 times in a row.

\subsection{Remaining Useful Life Prediction}

The second stage of the proposed framework is to predict the RUL as a percentage after the estimated FPC. Because the ground truth of the RUL is not given, the RUL labels after the estimated FPC should be defined and expressed as follows.
\begin{equation}
    \label{eq:RULLabels}
    {y_{L}}_j^i = \frac{\text{EOL}^i - j}{\text{EOL}^i - \text{FPC}^i}
\end{equation}
where ${y_{L}}_j^i$ represents the RUL label of $x_j^i$ after the FPC, EOL$^i$ and FPC$^i$ denote the EOL and FPC of $i$-th battery cell, respectively. In contrast to the conventional methods that solely relied on discharge capacity as a health indicator and predicted the discharge capacity values as the RUL prediction, we adopt a different approach in this work. The discharge capacity values often suddenly drop and the RUL is hard to be predicted accurately. Thus, in this work, we define the RUL as a percentage from the FPC to the EOL.

The RUL prediction model also utilizes a similar architecture to the 1D LSTM-based HS division model. Since the outputs of the RUL prediction model are the RUL as a percentage, we use ReLU activation function in the last FC layer. By applying ReLU activation, we enforce the predicted RUL to be within a valid range and avoid negative predictions. This adjustment in the activation function of the last FC layer aligns with the specific requirement of predicting RUL as a percentage. To train the RUL prediction model, we use mean absolute error (MAE) between the predicting RUL and the corresponding RUL labels.
\begin{equation}
    \label{eq:MAELoss}
    L_{MAE} = \frac{1}{\text{EOL}^i - \text{FPC}^i}\sum_{j=\text{FPC}^i}^{\text{EOL}^i} |{y_{L}}_j^i - {\hat{y_L}}_j^i |
\end{equation}
where ${y_{L}}_j^i$ and ${\hat{y_L}}_j^i$ represent the corresponding RUL label and RUL prediction of the $i$-th battery cell data, respectively.

\section{Experimental Results}
\label{sec:experiments}
\begin{figure}[!t]
\centering
\includegraphics[width=\columnwidth]{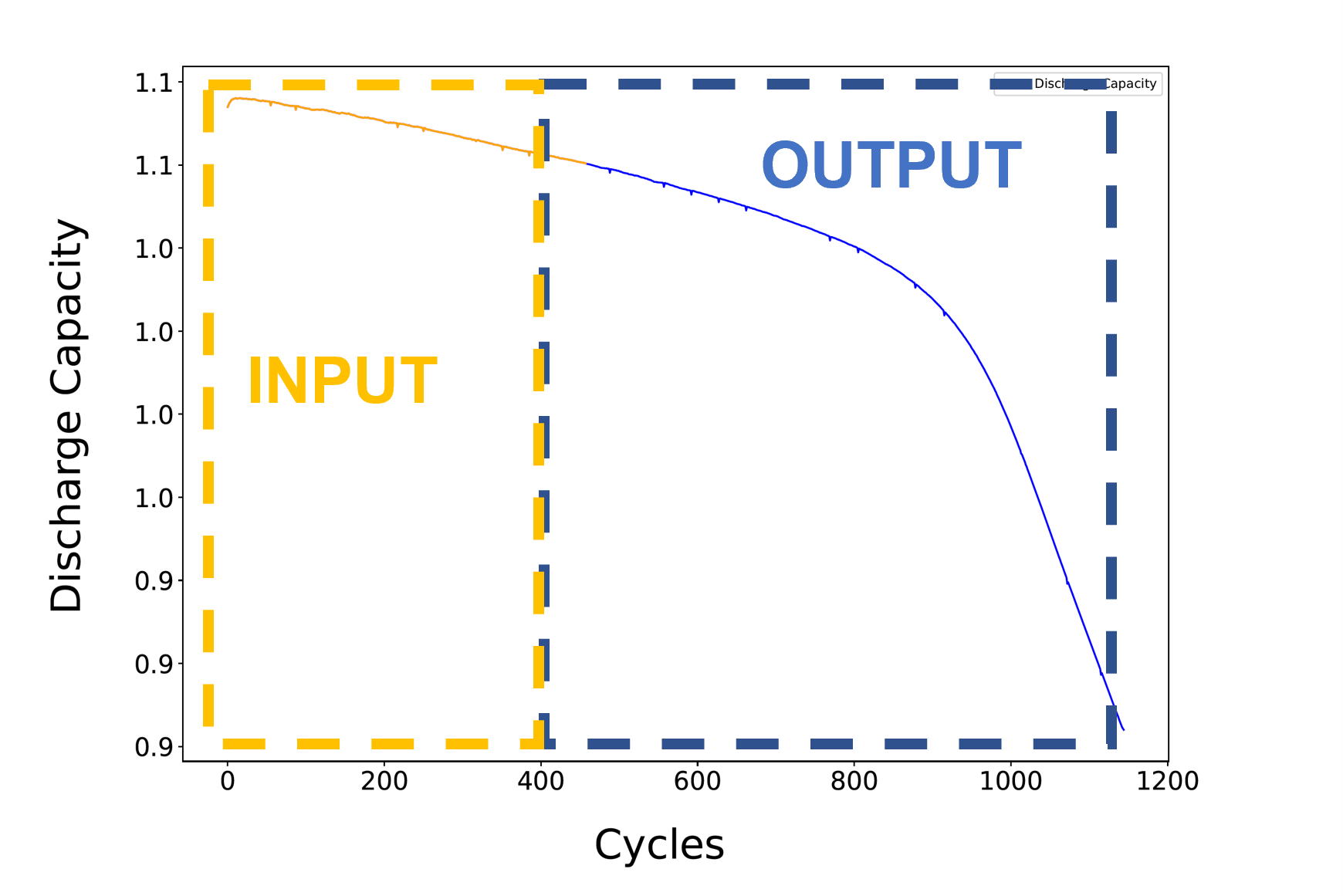}
\caption{The Conventional Scheme for RUL Prediction with Input and Output On The Same Battery Cell
}
\label{conventional}
\end{figure}

In this section, we present the experimental results of the proposed framework and provide a comprehensive comparison with conventional methods that directly use discharge capacity as sole input and assume prior knowledge of the EOL. \cref{conventional} shows the RUL prediction scheme of the conventional methods \cite{li2019remaining, zhou2020state}, where a specific percentage of the total cycles is given as input, and the remaining percentage is predicted as the output. Following the conventional methods \cite{li2019remaining, zhou2020state, tong2021early}, we used 40\% of the data from each battery cell data as the input and the remaining 60\% as the ground truth in the training dataset and as the output in the test dataset. In addition, we show the experimental results by varying the number of input features to show the effectiveness of utilizing different features.


\subsection{Implementation Details and Evaluation Metrics}
    The experiments were all implemented using Python scripts in the PyTorch framework. The input features consist of 7 variables, namely discharge capacity, internal resistance, charge capacity, average temperature, minimum and maximum temperature, and charge time. 
    We chose the Adam optimizer with a learning rate of 0.001, $\beta_1=0.9$, and $\beta_2=0.99$ to train the proposed models. The batch size was set to 8, and the training epochs were 100 using the early stopping with the patience of 20 epochs to prevent overfitting. 
    In the MIT dataset, We used 100 battery cell data for the training dataset and 24 battery cell data for the test dataset. 

    To evaluate and compare the proposed approach with conventional methods, we utilized three evaluation metrics: MAE, mean squared error (MSE), and mean absolute percentage error (MAPE), which are calculated as follows.
    \begin{equation}
        MSE^i = \frac{1}{N_C}\sum_{j}^{N_C} ({y_{L}}_j^i - {\hat{y_L}}_j^i)^2
        \label{eq:MSE}
    \end{equation}
    \begin{equation}
        MAPE^i = \frac{1}{N_C}\sum_{j}^{N_C} \frac{|{y_{L}}_j^i - {\hat{y_L}}_j^i|}{{y_{L}}_j^i}
        \label{eq:MAPE}
    \end{equation}
    where $N_C$ is the total number of cycles of the $i$-th battery cell. 
    
\subsection{Results and Analysis}

\begin{figure}[!t]
\centering
\includegraphics[width=\columnwidth]{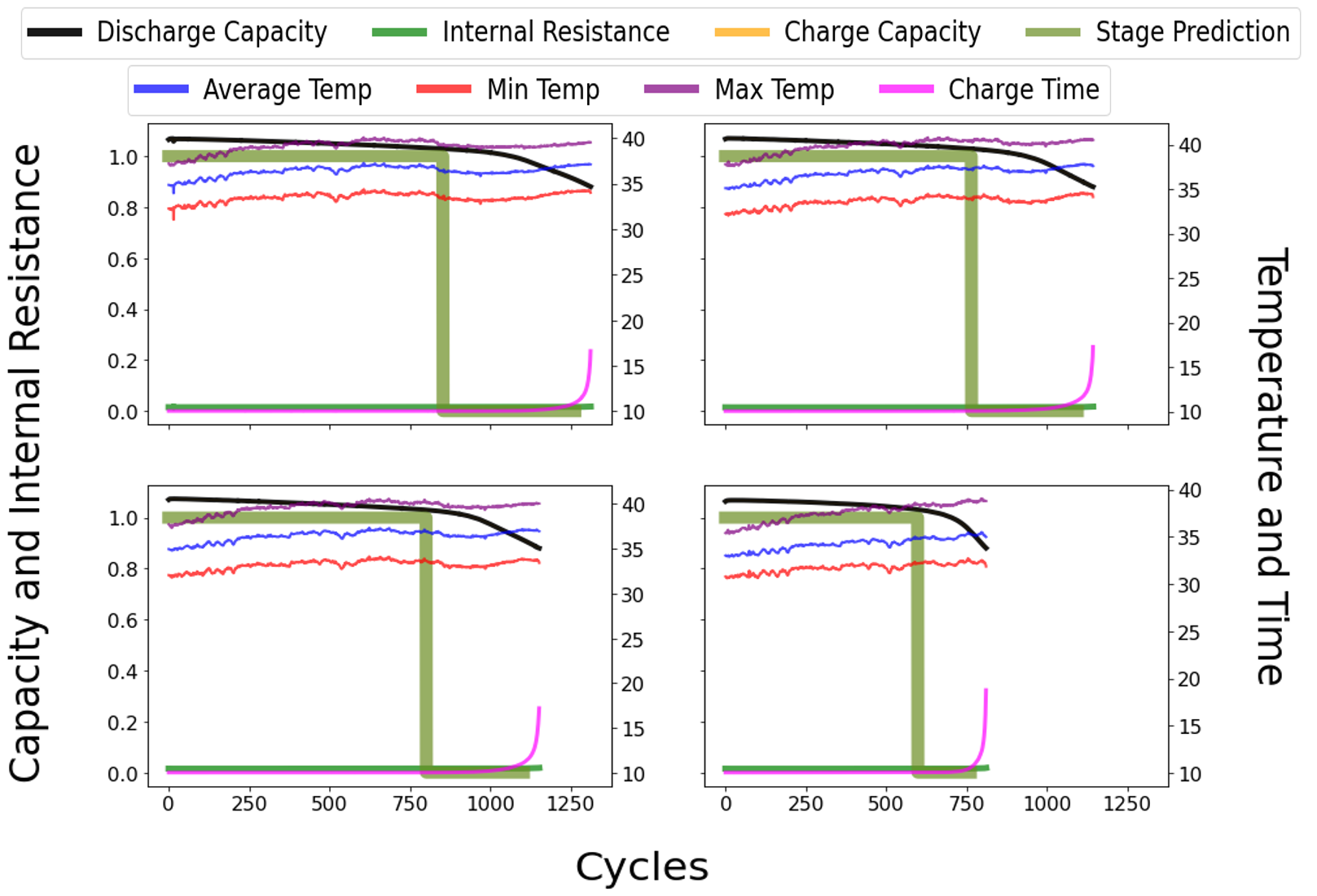}
\caption{First Predicted Cycle (FPC) Decision using as Input Seven Channels; Discharge Capacity, Internal Resistance, Charge Capacity, Average Temperature, Minimum and Maximum Temperature, and Charge Time. In the case of Battery Numbers \textbf{Upper Left} 101, \textbf{Upper Right} 102, \textbf{Lower Left} 103 and \textbf{Lower Right} 104}
\label{FPC_Testing}
\end{figure}

\cref{FPC_Testing} demonstrates example plots of the FPC decision with seven input feature data. It shows that the HS division classified the healthy state and unhealthy state clearly. Notably, the FPCs were determined at specific points where the degradation of discharge and charge capacity values began. our analysis revealed that, on average, the FPCs occurred at approximately 95\% of the total discharge capacity for any given battery used for testing and training data. 


\begin{figure}[!t]
\centering
\includegraphics[width =\linewidth]{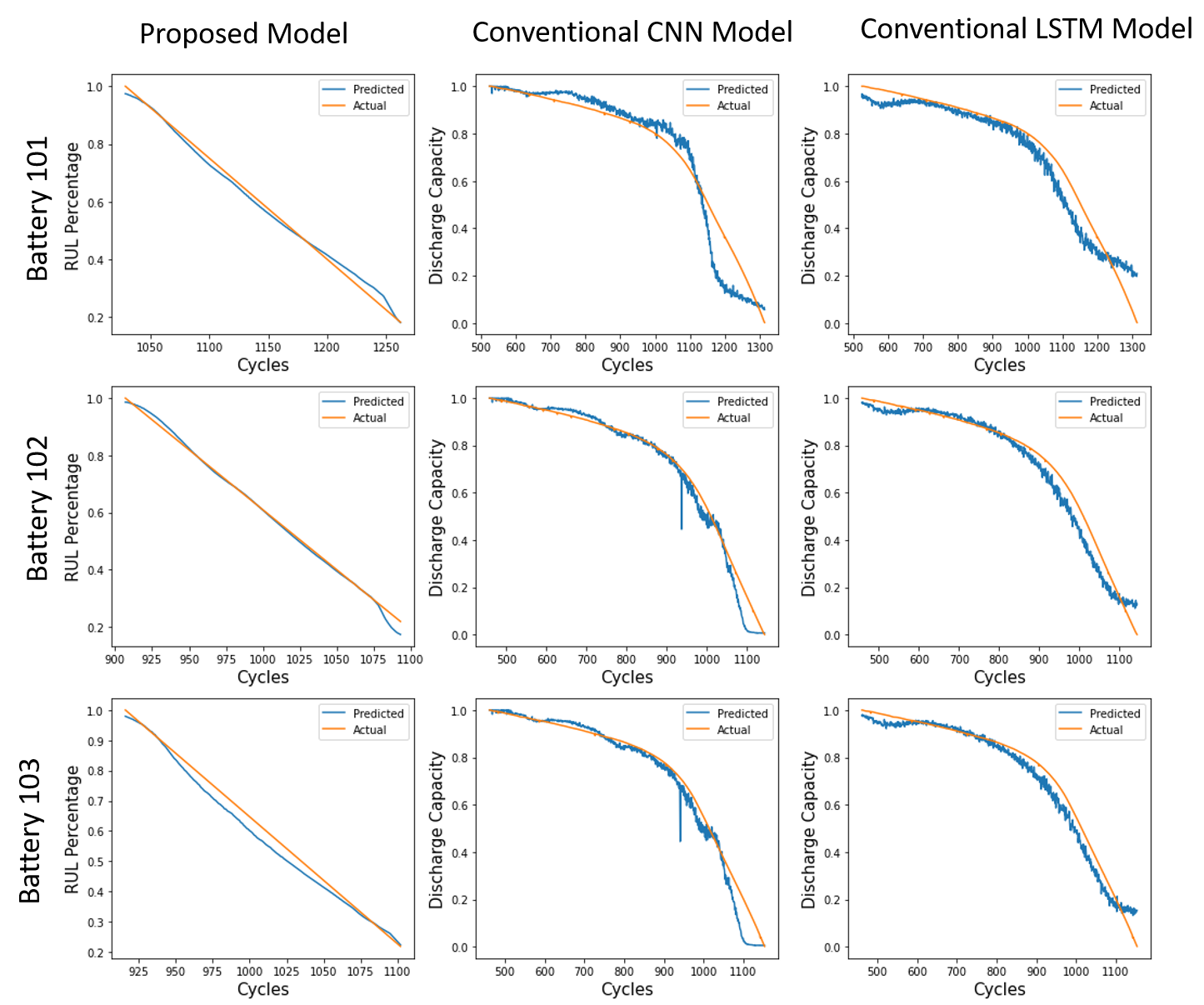}
\caption{Comparison of Our Approach With Conventional Approaches for Battery 101, 102 and 103. \textbf{First Column} Our Approach Estimation the Remaining Useful Life Percentage of the Batteries. \textbf{Second and Third Column} Conventional CNN \cite{zhou2020state} LSTM Model \cite{li2019remaining} of the Discharge Capacity of the Batteries.}
\label{comparison}
\end{figure}

\begin{table}[!t]
\caption{Comparison result of the RUL prediction models, where the proposed model is trained using four features}
\centering
\renewcommand{\arraystretch}{1.3}
\begin{tabular}{|c|c|c|c|} 
    \hline
    Method & MSE & MAE &  MAPE\\
    \hline
    Proposed  & \textbf{0.0035} & \textbf{0.0454} & \textbf{0.0882} \\
    \hline
    Zhou et al. \cite{zhou2020state} &0.0090  & 0.0522  & 0.0944  \\
    \hline
    Li et al. \cite{li2019remaining} &0.0054  & 0.0525  & 0.0902\\
    \hline   
\end{tabular}
\label{table:train_result}
\end{table}

\cref{comparison} shows examples of the RUL prediction results obtained by comparing the proposed method with conventional models \cite{zhou2020state, li2019remaining}. In the proposed method, RUL percentage is predicted after the FPC point whereas in conventional methods it is predicted after 40\% of the overall cycle data, making the starting points for RUL percentage prediction different in \cref{comparison}. While the conventional methods rely on prior knowledge of the EOL, the RUL predictions exhibit gaps with the ground truth and struggle to accurately predict sudden drops. On the other hand, the proposed method can predict smooth and stable RUL. In \cref{table:train_result}, a comprehensive comparison of the RUL prediction models is presented using three evaluation metrics. Notably, the proposed method outperformed the two conventional methods across all evaluation metrics, despite not requiring prior knowledge of the EOL.

\begin{table}[!t]
\caption{The RUL prediction results by changing the number of input features}
\centering
\renewcommand{\arraystretch}{1.3}
\begin{tabular}{|c|c|c|c|} 
    \hline
    \# of features & MSE & MAE &  MAPE\\
    \hline
    1  & 0.0133  & 0.0943  &  0.2477 \\
    \hline
    2  &  0.0062  & 0.0653  & 0.1199  \\
    \hline
    3  &  0.0065  & 0.0671  &  0.1154  \\
    \hline
    4  &  0.0035 & 0.0454  & 0.0882 \\
    \hline   
    7  &  0.0060  & 0.0606  & 0.1776  \\
    \hline   
\end{tabular}
\label{table:ablation}
\end{table}

We conducted experiments by adding input features in the following order: discharge capacity, charge capacity, internal resistance, charge time, average temperature, minimum temperature, and maximum temperature. Notably, the three temperature features were added simultaneously.
The results of the RUL prediction experiments varying the number of input features are summarized in \cref{table:ablation}. We observed a consistent trend of improved performance as the number of input features increased. However, interestingly, we noticed a slight degradation in performance when all the temperature features were included. This can be attributed to the fact that temperature, being a control variable, does not contribute significantly to the performance of RUL prediction. Thus, including all temperature features cannot provide substantial benefits.

\section{Conclusion}
\label{sec:conclusion}

In this paper, we proposed a novel two-stage early prediction framework of RUL for Lithium-ion batteries. The framework consists of the FPC decision stage and the RUL prediction stage. In the FPC decision stage, we classified the healthy and unhealthy states of the battery by training an LSTM (Long Short Term Memory) model. It helps to determine the FPC, where the transition from the healthy state to the unhealthy state occurs. In the RUL prediction stage, the proposed method utilized an LSTM-based RUL prediction model to estimate the RUL in percentage from the FPC to the EOL. Unlike conventional schemes that solely rely on discharge capacity as a health indicator, the proposed method considers multiple features. Experimental results demonstrated the effectiveness of the proposed framework. We compared it with conventional models that directly use discharge capacity as the sole input and assume prior knowledge of the EOL. The proposed method outperformed the conventional models across three evaluation metrics, even without requiring prior knowledge of the EOL.

For future work, we plan to apply this work to larger and more diverse public datasets to validate its generalizability and robustness. In addition, we will explore different neural network architectures, such as transformer \cite{chen2022transformer}, that could provide insights into capturing temporal dependencies and long-term patterns. Furthermore, We aim to delve deeper into the chemical properties of batteries to better understand how internal reactions impact charge degradation. Incorporating physics-based models and considering capacity-related physics could also improve the robustness and reliability of RUL predictions.



\section*{Acknowledgment}

This work was supported by Carl-Zeiss Stiftung under the Sustainable Embedded AI project (P2021-02-009) and by the European Union under the HumanE AI Network (H2020-ICT-2019-3 \#952026).

\newpage
\bibliographystyle{IEEEtran}
\bibliography{ref}

\end{document}